\documentclass[a4paper,10pt]{article}
\usepackage{times}
\usepackage{amsmath}
\usepackage{amssymb}
\usepackage[ansinew]{inputenc}
\usepackage{bm}
\usepackage{float}
\usepackage{url}
\usepackage{algorithm}
\usepackage{algorithmic}
\usepackage{graphicx}
\usepackage{subfigure}
\usepackage[small]{caption}
\usepackage[round, authoryear]{natbib}
\usepackage{color}
\usepackage{graphics}
\usepackage{graphicx}
\usepackage{multirow}

\newcommand{\latentForce}{u}
\newcommand{\LatentForce}{U}
\newcommand{\displacement}{y}
\newcommand{\Displacement}{Y}

\newcommand{\Sensitivity}{S}
\newcommand{\DecayRate}{B}
\newcommand{\DampingCoefficient}{C}
\newcommand{\Mass}{M}

\newcommand{\dif}[1]{\mathrm{d}#1}

\newcommand{\Complex}{\mathbb{C}}

\newcommand{\wfunc}[1]{\textnormal{w}(#1)}

\DeclareMathOperator{\vecO}{vec} 
\DeclareMathOperator{\cov}{cov} 
\newcommand{\boldK}{\mathbf{K}} 
\newcommand{\boldf}{\mathbf{f}} 
\newcommand{\boldA}{\mathbf{A}} 
\newcommand{\boldu}{\mathbf{u}} 
\newcommand{\boldy}{\mathbf{y}} 
\newcommand{\boldx}{\mathbf{x}} 
\newcommand{\eye}{\mathbf{I}}   
\newcommand{\boldX}{\mathbf{X}} 
\newcommand{\params}{\bm{\theta}} 
\newcommand{\gauss}{\mathcal{N}} 
\newcommand{\bolds}{\mathbf{s}}

\title{Linear Latent Force Models using Gaussian Processes}

\author{Mauricio A. \'Alvarez$^{\dagger,\ddagger}$, David Luengo$^\sharp$, Neil D. Lawrence$^{\star,\circ}$\\
{\small $\dagger$ \emph{School of Computer Science, University of Manchester, Manchester, UK M13 9PL.}}\\ 
{\small $\ddagger$ \emph{Faculty of Engineering, Universidad Tecnológica de Pereira, Colombia, 660003.}}\\
{\small $\sharp$ \emph{Dep. de la Teor\'ia de la Se\~nal y Comunicaciones, Universidad Carlos III de Madrid, 
28911 Legan\'es, Espa\~na.}}\\
{\small $\star$ \emph{School of Computer Science, University of Sheffield, Sheffield, UK S1 4DP.}}\\
{\small $\circ$ \emph{The Sheffield Institute for Translational Neuroscience, Sheffield, UK S10 2HQ.}}\\}
\date{}

\begin{document}
\maketitle

\begin{abstract}
Purely data driven approaches for machine learning present difficulties when data is scarce relative to the complexity of 
the model or when the model is forced to extrapolate. On the other hand, purely mechanistic approaches need to identify 
and specify all the interactions in the problem at hand (which may not be feasible) and still leave the issue of how to 
parameterize the system. In this paper, we present a hybrid approach using Gaussian processes and differential equations 
to combine data driven modelling with a physical model of the system. We show how different, physically-inspired, kernel 
functions can be developed through sensible, simple, mechanistic assumptions about the underlying system. The versatility 
of our approach is illustrated with three case studies from motion capture, computational biology and geostatistics.
\end{abstract}


\section{Introduction} \label{sec:intro}
Traditionally the main focus in machine learning has
been model generation through a \emph{data driven paradigm}. The usual
approach is to combine a data set with a (typically fairly flexible)
class of models and, through judicious use of regularization, make
predictions on previously unseen data. There are two key problems with
purely data driven approaches. Firstly, if data is scarce relative to
the complexity of the system we may be unable to make accurate
predictions on test data. Secondly, if the model is forced to
extrapolate, \emph{i.e.}\ make predictions in a regime in which data
has not yet been seen, performance can be poor. 

Purely \emph{mechanistic models}, \emph{i.e.}  models which are
inspired by the underlying physical knowledge of the system, are
common in many domains such as chemistry, systems biology, climate
modelling and geophysical sciences, \emph{etc.} They normally make use
of a fairly well characterized physical process that underpins the
system, often represented with a set of differential equations. The
purely mechanistic approach leaves us with a different set of problems
to those from the data driven approach. In particular, accurate
description of a complex system through a mechanistic modelling
paradigm may not be possible. Even if all the physical processes can
be adequately described, the resulting model could become extremely
complex. Identifying and specifying all the interactions might not be
feasible, and we would still be faced with the problem of identifying
the parameters of the system.

Despite these problems, physically well characterized models retain a
major advantage over purely data driven models. A mechanistic model
can enable accurate prediction even in regions where there is no
available training data. For example, Pioneer space probes can enter
different extra terrestrial orbits regardless of the availability of
data for these orbits.

Whilst data driven approaches do seem to avoid mechanistic assumptions
about the data, typically the regularization which is applied encodes
some kind of physical intuition, such as the smoothness of the
interpolant. This does reflect a weak underlying belief about the
mechanism that generated the data. In this sense the data driven
approach can be seen as \emph{weakly mechanistic} whereas models based
on more detailed mechanistic relationships could be seen as
\emph{strongly mechanistic}.

The observation that weak mechanistic assumptions underlie a data
driven model inspires our approach. We suggest a \emph{hybrid system}
which involves a (typically overly simplistic) mechanistic model of
the system. The key is to retain sufficient flexibility in our model
to be able to fit the system even when our mechanistic assumptions
are not rigorously fulfilled in practise. To illustrate the framework
we will start by considering dynamical systems as latent variable
models which incorporate ordinary differential equations. In this we
follow the work of \citet{Lawrence:gpsim2007a} and \citet{Gao:latent08} who encoded
a first order differential equation in a Gaussian process
(GP). However, their aim was to construct an accurate model of
transcriptional regulation, whereas ours is to make use of the
mechanistic model to incorporate salient characteristics of the data
(\emph{e.g.}\ in a mechanical system \emph{inertia}) without
necessarily associating the components of our mechanistic model with
actual physical components of the system. For example, for a human
motion capture dataset we develop a mechanistic model of motion
capture that does not exactly replicate the \emph{physics} of human
movement, but nevertheless captures salient features of the
movement. Having shown how linear dynamical systems can be
incorporated in a GP, we finally show how partial differential
equations can also be incorporated for modelling systems with multiple
inputs.

The paper is organized as follows. In section \ref{section:lvm:to:lfm} we motivate the latent force model using as an 
example a latent variable model. Section \ref{section:operational} employs a first order latent force model to describe
how the general framework can be used in practise. We then proceed to show three case studies. In section 
\ref{section:ode2} we use a latent force model based on a second order ordinary differential equation 
for characterizing motion capture datasets. Section \ref{Sec:Diffusion} presents a latent force model for spatio-temporal
domains applied to represent the development of Drosophila Melanogaster, and a latent force model inspired in a 
diffusion process to explain the behavior of pollutant metals in the Swiss Jura. Extensive related work is presented 
in section \ref{section:related:work}. Final conclusions are given in section \ref{section:conclusion}.

\section{From latent variables to latent functions}\label{section:lvm:to:lfm}

A key challenge in combining the mechanistic and data-driven approaches is how to incorporate the model flexibility associated with the data-driven approach within the mechanism. We choose to do this through latent variables, more precisely latent functions: unobserved functions from the system. To see how this is possible we first introduce some well known data driven models from a mechanistic latent-variable perspective. 

Let us assume we wish to summarize a high dimensional data set with a
reduced dimensional representation. For example, if our data consists
of $N$ points in a $D$ dimensional space we might seek a linear
relationship between the data,
$\mathbf{\Displacement}=[\mathbf{y}_1,\ldots,
\mathbf{y}_D]\in\mathbb{R}^{N\times D}$ with
$\mathbf{\displacement}_d\in\mathbb{R}^{N\times 1}$, and a reduced
dimensional representation,
$\mathbf{\LatentForce}=[\mathbf{\latentForce}_1, \ldots,
\mathbf{\latentForce}_Q] \in \mathbb{R}^{N\times Q}$ with
$\mathbf{\latentForce}_q\in\mathbb{R}^{N\times 1}$, where $Q <
D$. From a probabilistic perspective this involves an assumption that
we can represent the data as
\begin{align}
  \mathbf{\Displacement} & = \mathbf{\LatentForce} \mathbf{W}^{\top} +
  \mathbf{E},
\label{eq:linDimRed}
\end{align}
where $\mathbf{E}=[\mathbf{e}_1, \ldots, \mathbf{e}_D]$ is a
matrix-variate Gaussian noise: each column,
$\mathbf{e}_{d}\in\mathbb{R}^{N\times 1}$ ($1 \le d \le D$), is a
multi-variate Gaussian with zero mean and covariance
$\boldsymbol{\Sigma}$, this is, $\mathbf{e}_{d} \sim \mathcal{N}\left(
  \mathbf{0}, \boldsymbol{\Sigma}_d\right)$. The usual approach, as
undertaken in factor analysis and principal component analysis (PCA),
to dealing with the unknown latent variables in this model is to
integrate out $\mathbf{\LatentForce}$ under a Gaussian prior and
optimize with respect to $\mathbf{W} \in \mathbb{R}^{D \times Q}$
(although it turns out that for a non-linear variant of the model it
can be convenient to do this the other way around, see for example
\citet{Lawrence:pnpca05}). If the data has a temporal nature, then the
Gaussian prior in the latent space could express a relationship
between the rows of $\mathbf{\LatentForce}$,
$\mathbf{\latentForce}_{t_n}=
\bm{\Gamma}\mathbf{\latentForce}_{t_{n-1}}+\boldsymbol{\eta}$, where
$\bm{\Gamma}$ is a transformation matrix, $\boldsymbol{\eta}$ is a
Gaussian random noise and $\mathbf{\latentForce}_{t_n}$ is the $n$-th
row of $\mathbf{\LatentForce}$, which we associate with time
$t_n$. This is known as the \emph{Kalman filter/smoother}. Normally
the times, $t_n$, are taken to be equally spaced, but more generally
we can consider a joint distribution for
$p\left(\mathbf{\LatentForce}|\mathbf{t}\right)$, for a vector of time
inputs $\mathbf{t}=\left[t_1 \dots t_N\right]^{\top}$, which has the
form of a Gaussian process,
\begin{align}
p\left(\mathbf{\LatentForce}|\mathbf{t}\right) &= \prod_{q=1}^Q\mathcal{N}\left(\mathbf{\latentForce}_{q}|\mathbf{0},
  \mathbf{K}_{\mathbf{\latentForce}_{q},\mathbf{\latentForce}_{q}}\right), 
\end{align}
where we have assumed zero mean and independence across the $Q$
dimensions of the latent space. The GP makes explicit the fact that
the latent variables are functions,
$\left\{\latentForce_q(t)\right\}_{q=1}^Q$, and we have now described
them with a process prior. The elements of the vector
$\mathbf{\latentForce}_{q}=[\latentForce_q(t_1),\ldots,
\latentForce_q(t_N)]^{\top}$, represents the values of the function
for the $q$-th dimension at the times given by $\mathbf{t}$. The
matrix
$\mathbf{K}_{\mathbf{\latentForce}_{q},\mathbf{\latentForce}_{q}}$ is
the covariance function associated to $\latentForce_q(t)$ computed at
the times given in $\mathbf{t}$.

Such a GP can be readily implemented. Given the covariance functions
for $\left\{\latentForce_q(t)\right\}_{q=1}^Q$ the implied covariance
functions for $\left\{\displacement_d(t)\right\}_{d=1}^D$ are
straightforward to derive. In \citet{Teh:semiparametric05} this is
known as a semi-parametric latent factor model (SLFM), although their
main focus is not the temporal case. If the latent functions $u_q(t)$
share the same covariance, but are sampled independently, this is
known as the multi-task Gaussian process prediction model (MTGP)
\citep{Bonilla:multi07} with a similar model introduced in
\citet{Rogers:towards08}. Historically the Kalman filter approach has
been preferred, perhaps because of its linear computational complexity
in $N$. However, recent advances in sparse approximations have made
the general GP framework practical (see \citet{Quinonero:unifying05}
for a review).

So far the model described relies on the latent variables to provide
the dynamic information. Our main contribution is to include a further
dynamical system with a \emph{mechanistic} inspiration. We will make use of 
a mechanical analogy to introduce it. Consider
the following physical interpretation of (\ref{eq:linDimRed}): the
latent functions, $\latentForce_q(t)$, are $Q$ forces and we observe
the displacement of $D$ springs, $y_d(t)$, to the forces. Then we can
reinterpret \eqref{eq:linDimRed} as the force balance equation,
$\mathbf{\Displacement} \mathbf{\DecayRate} = \mathbf{\LatentForce}
\mathbf{\Sensitivity}^{\top} + \widetilde{\mathbf{E}}$. Here we have
assumed that the forces are acting, for example, through levers, so
that we have a matrix of sensitivities,
$\mathbf{\Sensitivity}\in\mathbb{R}^{D \times Q}$, and a diagonal
matrix of spring constants, $\mathbf{\DecayRate}\in\mathbb{R}^{D\times
  D}$, with elements $\{B_d\}_{d=1}^D$.  The original model is
recovered by setting
$\mathbf{W}^{\top}=\mathbf{S}^{\top}\mathbf{B}^{-1}$ and
$\tilde{\mathbf{e}}_{d} \sim \mathcal{N}\left(
  \mathbf{0},\mathbf{B}^{\top} \boldsymbol{\Sigma_d} \mathbf{B}
\right)$. With appropriate choice of latent density and noise model
this physical model underlies the Kalman filter, PCA, independent
component analysis and the multioutput Gaussian process models we
mentioned above. The use of latent variables means that despite this
strong physical constraint these models are still powerful enough to
be applied to a range of real world data sets. We will retain this
flexibility by maintaining the latent variables at the heart of the
system, but introduce a more realistic system by extending the
underlying physical model. Let us assume that the springs are acting in
parallel with dampers and that the system has mass, allowing us to
write,
\begin{align}\label{eq:lfm}
\ddot{\mathbf{\Displacement}}\mathbf{\Mass}+\dot{\mathbf{\Displacement}}
\mathbf{\DampingCoefficient} + \mathbf{\Displacement}\mathbf{\DecayRate}& = \mathbf{\LatentForce}\mathbf{\Sensitivity} 
+ \widehat{\mathbf{E}},
\end{align}
where $\mathbf{\Mass}$ and $\mathbf{\DampingCoefficient}$ are diagonal
matrices of masses, $\{M_d\}_{d=1}^D$, and damping coefficients,
$\{C_d\}_{d=1}^D$, respectively, $\dot{\mathbf{\Displacement}}$ is the
first derivative of $\mathbf{\Displacement}$ with respect to time
(with entries $\{\dot{y}_d(t_n)\}$ for $d=1, \ldots D$ and $n=1,\dots,
N$), $\ddot{\mathbf{\Displacement}}$ is the second derivative of
$\mathbf{\Displacement}$ with respect to time (with entries
$\{\ddot{y}_d(t_n)\}$ for $d=1, \ldots D$ and $n=1,\dots, N$) and
$\widehat{\mathbf{E}}$ is once again a matrix-variate Gaussian
noise. Equation \eqref{eq:lfm} specifies a particular type of
interaction between the outputs $\mathbf{\Displacement}$ and the set
of latent functions $\mathbf{\LatentForce}$, namely, that a weighted
sum of the second derivative for $y_d(t)$, $\ddot{y}_d(t)$, the first
derivative for $y_d(t)$, $\dot{y}_d(t)$, and $y_d(t)$ is equal to the
weighted sum of functions $\{u_q(t)\}_{q=1}^Q$ plus a random
noise. The second order mechanical system that this model describes
will exhibit several characteristics which are impossible to represent
in the simpler latent variable model given by (\ref{eq:linDimRed}),
such as inertia and resonance. This model is not only appropriate for
data from mechanical systems. There are many analogous systems which
can also be represented by second order differential equations, for
example Resistor-Inductor-Capacitor circuits. A unifying
characteristic for all these models is that the system is being forced
by latent functions,
$\left\{\latentForce_q(t)\right\}_{q=1}^{Q}$. Hence, we refer to them
as \emph{latent force models} (LFMs). This is our general framework:
combine a physical system with a probabilistic prior over some latent
variable. 

One analogy for our model comes through puppetry. A marionette is a
representation of a human (or animal) controlled by a limited number
of inputs through strings (or rods) attached to the character. In a puppet show these inputs are the unobserved latent functions. Human motion is a high dimensional data set. A skilled puppeteer with a well designed puppet can create a realistic representation of human movement through judicious use of the strings  

\section{Latent Force Models in Practise}\label{section:operational}

In the last section we provided a general description of the latent
force model idea and commented how it compares to previous models in
the machine learning and statistics literature. In this section we
specify the operational procedure to obtain the Gaussian process model
associated to the outputs and different aspects involved in the
inference process.  First, we illustrate the procedure using a
first-order latent force model, for which we assume there are no
masses associated to the outputs and the damper constants are equal to
one. Then we specify the inference procedure, which involves
maximization of the marginal likelihood for estimating
hyperparameters. Next we generalize the operational procedure for
latent force models of higher order and multidimensional inputs and
finally we review some efficient approximations to reduce
computational complexity.

\subsection{First-order Latent Force Model}\label{subsection:first:lfm}
Assume a simplified latent force model, for which only the first
derivative of the outputs is included. This is a particular case of
equation \eqref{eq:lfm}, with masses equal to zero and damper
constants equal to one. With these assumptions, equation
\eqref{eq:lfm} can be written as
\begin{align}\label{eq:sim}
\dot{\mathbf{\Displacement}} + \mathbf{\Displacement}\mathbf{\DecayRate} 
&= \mathbf{\LatentForce}\mathbf{\Sensitivity} + \widehat{\mathbf{E}}.
\end{align}
Individual elements in equation \eqref{eq:sim} follow
\begin{align}\label{Eq:GP-SIM:Differential_Equation}
\frac{\text{d}y_d(t)}{\text{d}t} + B_d y_d(t)& = \sum_{q=1}^{Q}{S_{d,q} u_q(t)} + \hat{e}_d(t).
\end{align}
Given the parameters $\{B_d\}_{d=1}^D$ and
$\{S_{d,q}\}_{d=1,q=1}^{D,Q}$, the uncertainty in the outputs is given
by the uncertainty coming from the set of functions
$\{u_q(t)\}_{q=1}^Q$ and the noise $\hat{e}_d(t)$. Strictly speaking,
this equation belongs to a more general set of equations known as
\emph{stochastic differential equations} (SDE) that are usually solved
using special techniques from stochastic calculus
\citep{Oksendal:SDEs:2003}.  The representation used in equation
\eqref{Eq:GP-SIM:Differential_Equation} is more common in physics,
where it receives the name of \emph{Langevin equations}
\citep{Reichl:statsPhysics:1998}. For the simpler equation
\eqref{Eq:GP-SIM:Differential_Equation}, the solution is found using
standard calculus techniques and is given by
\begin{align}\label{eq:sol:ode:sim}
y_d(t)&= y_d(t_0)e^{-B_dt} + \sum_{q=1\normalfont}^{Q}S_{d,q}{\mathcal{G}_{d}[u_q](t)} + \mathcal{G}_{d}[\hat{e}_d](t),
\end{align}
where $y_d(t_0)$ correspond to the value of $y_d(t)$ for $t=t_0$ (or
the initial condition) and $\mathcal{G}_d$ is a linear integral
operator that follows
\begin{align*}
\mathcal{G}_{d}[v](t) &= f_d(t,v(t))= \int_{0}^{t}{e^{-B_d(t-\tau)} v(\tau) \text{d}\tau}.
\end{align*}
Our noise model $\mathcal{G}_{d}[\hat{e}_d](t)$ has a particular form
depending on the linear operator $\mathcal{G}_d$.  For example, for
the equation in \eqref{eq:sol:ode:sim} and assuming a white noise
process prior for $e_d(t)$, it can be shown that the process
$\mathcal{G}_{d}[\hat{e}_d](t)$ corresponds to the Ornstein-Uhlenbeck
(OU) process \citep{Reichl:statsPhysics:1998}. In what follows, we will
allow the noise model to be a more general process and we denote it by
$w_d(t)$. Without loss of generality, we also assume that the initial
conditions $\{y_d(t_0)\}_{d=1}^D$ are zero, so that we can write again
equation \eqref{eq:sol:ode:sim} as
\begin{align}\label{eq:sol:ode:sim:2}
y_d(t)&= \sum_{q=1\normalfont}^{Q}S_{d,q}{\mathcal{G}_{d}[u_q](t)} + w_d(t).
\end{align}
We assume that the latent functions $\{u_q(t)\}_{q=1}^Q$ are
independent and each of them follows a Gaussian process prior, this
is, $\latentForce_q(t) \sim\mathcal{GP}(0,
k_{u_q,u_q}(t,t'))$.\footnote{We can allow a mean prior different from
  zero.} Due to the linearity of $\mathcal{G}_d$, $\{y_d(t)\}_{d=1}^D$
correspond to a Gaussian process with covariances
$k_{y_d,y_{d'}}(t,t')=\cov[y_d(t), y_{d'}(t')]$ given by
\begin{align*}
  \cov[f_d(t), f_{d'}(t')] + \cov[w_d(t), w_{d'}(t')]\delta_{d,d'},
\end{align*}
where $\delta_{d,d'}$ corresponds to the Kronecker delta and
$\cov[f_d(t), f_{d'}(t')]$ is given by
\begin{align*}
  \sum_{q=1}^QS_{d,q}S_{d',q}\cov[f^q_d(t), f^q_{d'}(t')],
\end{align*}
where we use $f^q_d(t)$ as a shorthand for $f_d(t, u_q(t))$.
Furthermore, for the latent force model in equation \eqref{eq:sim},
the covariance $\cov[f^q_d(t), f^q_{d'}(t')]$ is equal to
\begin{align}\label{eq:kfdfd}
\int_{0}^{t}e^{-B_d(t-\tau)}\int_{0}^{t'}e^{-B_{d'}(t'-\tau')}k_{u_q,u_q}(\tau,\tau')\text{d}\tau'\text{d}\tau.
\end{align}
Notice from the equation above that the covariance between $f^q_d(t)$
and $f^q_{d'}(t')$ depends on the covariance
$k_{u_q,u_q}(\tau,\tau')$. We alternatively denote $\cov[f_d(t),
f_{d'}(t')]$ as $k_{f_d,f_{d'}}(t,t')$ and $\cov[f^q_d(t),
f^q_{d'}(t')]$ as $k_{f^q_d,f^q_{d'}}(t,t')$. The form for the
covariance $k_{u_q,u_q}(t,t')$ is such that we can solve both
integrals in equation \eqref{eq:kfdfd} and find an analytical
expression for the covariance $k_{f_d,f_{d'}}(t,t')$. In the rest of
the paper, we assume the covariance for each latent force
$\latentForce_q(t)$ follows the squared-exponential (SE) form
\citep{Rasmussen:book06}
\begin{align}\label{eq:cov:se}
k_{u_q,u_q}(t,t') = \exp \left(-\frac{(t-t')^ 2}{\ell_q^2}\right),
\end{align}
where $\ell_q$ is known as the length-scale. We can compute the
covariance $k_{f^q_d,f^q_{d'}}(t,t')$ obtaining
\citep{Lawrence:gpsim2007a}
\begin{align}
k_{f^q_d,f^q_{d'}}(t, t') = \frac{\sqrt{\pi} \ell_q}{2} [h_{d',d}(t', t) + h_{d,d'}(t, t')],\label{eq:cov:sim:basic}
\end{align}
where
\begin{align}
h_{d',d}(t', t) = \frac{\exp(\nu_{q,d'}^2)}{B_d+B_{d'}}& \exp(-B_{d'} t') \Bigg\{ \exp(B_{d'} t)  
\bigg[ \text{erf}\left(\frac{t'-t}{\ell_q}-\nu_{q,d'}\right) + \text{erf}\left(\frac{t}{\ell_q}+\nu_{q,d'}
\right) \bigg]\nonumber \\
&- \exp(-B_d t)\left[\text{erf}\left(\frac{t'}{\ell_q}-\nu_{q,d'}\right)+\text{erf}(\nu_{q,d'})\right] \Bigg\}, 
\end{align}
where $\text{erf}(x)$ is the real valued error function, $\text{erf}(x) = \frac{2}{\sqrt{\pi}}\int_{0}^{x}{\exp(-y^2)
\text{d}y}$, and $\nu_{q,d} = \ell_q B_d/2$. The covariance function in equation \eqref{eq:cov:sim:basic} is nonstationary.
For the stationary regime, the covariance function can be obtained by writing $t'=t+\tau$ and taking the limit as $t$ tends
to infinity. This is, 
$k^{\text{STAT}}_{f^q_d,f^q_{d'}}(\tau)=\lim_{t\rightarrow \infty}k_{f^q_d,f^q_{d'}}(t, t+\tau)$. The stationary 
covariance could also be obtained making use of the power spectral density for the stationary processes $u_q(t)$, 
$U_q(\omega)$ and the transfer function $H_d(\omega)$ associated to $h_d(t-s)=e^{-B_d(t-s)}$, the impulse response of the 
first order dynamical system. Then applying the convolution property of the Fourier transform to obtain the power spectral 
density of $f^q_d(t)$, $F^q_d(\omega)$, and finally using the \emph{Wiener-Khinchin theorem} to find the solution for 
$f^q_d(t)$ \citep{Shanmugan:randomSignals:88}. 

As we will see in the following section, for computing the posterior distribution for $\{u_q(t)\}_{q=1}^Q$, we need the 
cross-covariance between the output $y_d(t)$ and the latent force $u_q(t)$. Due to the independence between $u_q(t)$ and 
$w_d(t)$, the covariance reduces to $k_{f_d,u_q}(t,t')$, given by  
\begin{align}\label{eq:cross:output:latent}
k_{f_d, u_q}(t, t') & =\frac{\sqrt{\pi} \ell_q S_{d,q}}{2} \exp(\nu_{q,d}^2) \exp(-B_d(t-t'))
\bigg[\text{erf}\left(\frac{t-t'}{\ell_q}-\nu_{q,d}\right) + \text{erf}\left(\frac{t'}{\ell_q}+\nu_{q,d}\right)
\bigg].  
\end{align}
\subsection{Hyperparameter learning}\label{subsection:hyperparameter:learning}
We have implicitly marginalized out the effect of the latent forces using the Gaussian process prior for 
$\{u_q(t)\}_{q=1}^Q$ and the covariance for the outputs 
after marginalization is given by $k_{y_d,y_{d'}}(t,t')$. Given a set of inputs $\mathbf{t}$ and the
parameters of the covariance function,\footnote{Also known as hyperparameters.} $\bm{\theta}=(\{B_d\}_{d=1}^D,
\{S_{d,q}\}_{d=1, q=1}^{D,Q}, \{\ell_q\}_{q=1}^Q)$, the marginal likelihood for the outputs can be 
written as 
\begin{align}\label{eq:marginal:likelihood}
p(\mathbf{\displacement}|\mathbf{t},\bm{\theta})&=\mathcal{N}(\mathbf{\displacement}|\mathbf{0}, 
\boldK_{\boldf,\boldf} + \bm{\Sigma}),
\end{align}
where $\mathbf{\displacement}=\vecO{\mathbf{\Displacement}}$,\footnote{$\mathbf{x}=\vecO{\boldX}$ is the vectorization 
operator that transforms the matrix $\boldX$ into a vector $\mathbf{x}$. The vector is obtained by stacking the columns of 
the matrix.} $\boldK_{\boldf,\boldf}\in \mathbb{R}^{ND\times ND}$ with each element given by 
$\cov[f_d(t_n), f_{d'}(t'_{n'})]$ for $n=1, \ldots, N$ and $n'=1, \ldots, N$ and $\boldsymbol{\Sigma}$ represents the 
covariance associated with the independent processes $w_d(t)$. In general, the vector of parameters $\bm{\theta}$ is 
unknown, so we estimate it by maximizing the marginal likelihood. 

For clarity, we assumed that all outputs are evaluated at the same set of inputs $\mathbf{t}$. However, due to the 
flexibility provided by the Gaussian process formulation, each output can have associated a specific set of inputs, this 
is $\mathbf{t}_d=[t^d_1, \ldots, t^d_{N_d}]$.   

Prediction for a set of input test $\mathbf{t}_*$ is done using standard Gaussian process regression techniques. 
The predictive distribution is given by
\begin{align}\label{eq:predictive:distribution}
p(\mathbf{\displacement}_*|\mathbf{\displacement},\mathbf{t},\bm{\theta})&=\mathcal{N}(\mathbf{\displacement}_*|\bm{\mu}_*, 
\boldK_{\mathbf{\displacement}_*, \mathbf{\displacement}_*}),
\end{align}
with 
\begin{align*}
\bm{\mu}_*&=\boldK_{\boldf_*, \boldf}\left(\boldK_{\boldf, \boldf}+\bm{\Sigma}\right)^{-1}\mathbf{y},\\
\boldK_{\mathbf{\displacement}_*, \mathbf{\displacement}_*} &= \boldK_{\boldf_*, \boldf_*}-\boldK_{\boldf_*, \boldf}\left(
\boldK_{\boldf, \boldf}+\bm{\Sigma}\right)^{-1}\boldK^{\top}_{\boldf_*, \boldf}+\bm{\Sigma}_*,
\end{align*}
where we have used $\boldK_{\boldf_*, \boldf_*}$ to represent the evaluation of $\boldK_{\boldf, \boldf}$ at the input set
$\mathbf{t}_*$. The same meaning is given to the covariance matrix $\boldK_{\boldf_*, \boldf}$. 

As part of the inference process, we are also interested in the posterior distribution for the set of latent forces,
\begin{align}\label{eq:posterior:latent:forces}
p(\boldu|\mathbf{y},\mathbf{t},\bm{\theta})&=\mathcal{N}(\boldu|\bm{\mu}_{\bm{\boldu}|\mathbf{\displacement}}, 
\boldK_{\boldu| \mathbf{\displacement}}),
\end{align}
with 
\begin{align*}
\bm{\mu}_{\bm{\boldu}|\mathbf{\displacement}}&=\boldK^{\top}_{\boldf, \boldu}\left(\boldK_{\boldf, \boldf}+\bm{\Sigma}\right)^{-1}
\mathbf{y},\\
\boldK_{\bm{\boldu}|\mathbf{\displacement}} &= \boldK_{\boldu, \boldu}-\boldK_{\boldf, \boldu}^{\top}\left(
\boldK_{\boldf, \boldf}+\bm{\Sigma}\right)^{-1}\boldK_{\boldf, \boldu},
\end{align*}
where $\boldu=\vecO{\mathbf{\LatentForce}}$, $\boldK_{\boldu, \boldu}$ is a block-diagonal matrix with blocks given by 
$\boldK_{\boldu_q, \boldu_q}$. In turn, the elements of $\boldK_{\boldu_q, \boldu_q}$ are given by $k_{u_q,u_q}(t,t')$ 
in equation \eqref{eq:cov:se}, for $\{t_n\}_{n=1}^N$. Also $\boldK_{\boldf, \boldu}$ is a matrix with blocks 
$\boldK_{\boldf_d, \boldu_q}$, where $\boldK_{\boldf_d, \boldu_q}$ has entries given by $k_{f_d,u_q}(t,t')$ in equation
\eqref{eq:cross:output:latent}.
\subsection{Higher-order Latent Force Models}
In general, a latent force model of order $M$ can be described by the following equation
\begin{align}\label{eq:lfm:higher:order}
\sum_{m=0}^M\mathcal{D}^{m}[\mathbf{\Displacement}]\boldA_m& = \mathbf{\LatentForce}\mathbf{\Sensitivity}^{\top} + 
\widehat{\mathbf{E}},
\end{align}
where $\mathcal{D}^m$ is a linear differential operator such that $\mathcal{D}^m[\mathbf{\Displacement}]$ is a matrix with 
elements given by $\mathcal{D}^{m}y_d(t)=\frac{\dif^m y_d(t)}{\dif^m t}$ and $\boldA_m$ is a diagonal matrix with elements 
$A_{m,d}$ that weights the contribution of $\mathcal{D}^{m}y_d$. 

We follow the same procedure described in section \ref{subsection:first:lfm} for the model in equation 
\eqref{eq:lfm:higher:order} with $M=1$. Each element in expression \eqref{eq:lfm:higher:order} can be written as
\begin{align}\label{eq:lfm:higher:order:individual}
\mathcal{D}^{M}_0\displacement_d&=\sum_{m=0}^MA_{m,d}\mathcal{D}^{m}\displacement_d(t)
= \sum_{q=1}^{Q}{S_{d,q} u_q(t)} + \hat{e}_d(t),
\end{align}
where we have introduced a new operator $\mathcal{D}^{M}_0$ that is equivalent to apply the weighted sum of operators 
$\mathcal{D}^{m}$. For a homogeneous differential equation in \eqref{eq:lfm:higher:order:individual}, 
this is $u_q(t)=0$ for $q=1,\dots,Q$ and $e_d(t)=0$, and a particular set of initial conditions 
$\{\mathcal{D}^my_d(t_0)\}_{m=0}^{M-1}$, it is possible to find a linear integral operator $\mathcal{G}_d$
associated to $\mathcal{D}^{M}_0$ that can be used to solve the non-homogeneous differential equation. The linear integral
operator is defined as     
\begin{align}\label{eq:convolution:operator}
\mathcal{G}_{d}[v](t) &= f_d(t,v(t))= \int_{\mathcal{T}}G_d(t,\tau) v(\tau) \text{d}\tau, 
\end{align}
where $G_d(t,s)$ is known as the Green's function associated to the differential operator $\mathcal{D}^{M}_0$, $v(t)$ 
is the input function for the non-homogeneous differential equation and $\mathcal{T}$ is the input domain. The particular 
relation between the differential operator and the Green's function is given by 
\begin{align}\label{eq:diff:oper:greens}
\mathcal{D}^M_0[G_d(t,s)]&=\delta(t-s),  
\end{align}
with $s$ fixed, $G_d(t,s)$ a fundamental solution that satisfies the initial conditions and $\delta(t-s)$ the Dirac delta\footnote{We have used the same notation for the Kronecker delta and the Dirac delta. The particular meaning should be 
understood from the context.} function \citep{Griffel:functionalBook:2002}. Strictly speaking, the differential operator
in equation \eqref{eq:diff:oper:greens} is the adjoint for the differential operator appearing in equation \eqref{eq:lfm:higher:order:individual}. For a more rigorous introduction to Green's functions applied to differential equations, the interested reader is referred to \citet{Roach:greensBook:1982}. In the signal processing and control theory literature, the Green's
function is known as the impulse response of the system. Following the general latent force model framework, we write the 
outputs as
\begin{align}\label{eq:sol:ode:glfm}
y_d(t)&= \sum_{q=1\normalfont}^{Q}S_{d,q}{\mathcal{G}_{d}[u_q](t)} + w_d(t),
\end{align}
where $w_d(t)$ is again an independent process associated to each output. We assume once more that the latent forces
follow independent Gaussian process priors with zero mean and covariance $k_{u_q,u_q}(t,t')$. The covariance for the 
outputs $k_{y_d,y_{d'}}(t,t')$ is given by $k_{f_d,f_{d'}}(t,t') + k_{w_d,w_{d'}}(t,t')\delta_{d,d'}$, with $k_{f_d,f_{d'}}(t,t')$ 
equal to
\begin{align}\label{eq:cov:ode:glfm}
\sum_{q=1}^Q S_{d,q}S_{d',q}k_{f^q_d,f^q_{d'}}(t,t'),
\end{align}
and $k_{f^q_d,f^q_{d'}}(t,t')$ following
\begin{align}\label{eq:cov:ode:glfm:individual}
\int_{\mathcal{T}}\int_{\mathcal{T}'}G_d(t-\tau)G_{d'}(t'-\tau')k_{u_q,u_q}(\tau,\tau')\dif{\tau'}\dif\tau.
\end{align}
Learning and inference for the higher-order latent force model is done as explained in subsection 
\ref{subsection:hyperparameter:learning}. The Green's function is described by a parameter vector
$\bm{\psi}_d$ and with the length-scales $\{\ell_q\}_{q=1}^Q$ describing the latent GPs, the vector of hyperparameters
is given by $\bm{\theta}=\{\{\bm{\psi}_d\}_{d=1}^D, \{S_{d,q}\}_{d=1,q=1}^{D,Q},\{\ell_q\}_{q=1}^Q\}$. The parameter vector 
$\bm{\theta}$ is estimated by maximizing the logarithm of the marginal likelihood in equation 
\eqref{eq:marginal:likelihood}, where the elements of the matrix $\boldK_{\boldf,\boldf}$ are computed using expression \eqref{eq:cov:ode:glfm} with $k_{f^q_d,f^q_{d'}}(t,t')$ given by \eqref{eq:cov:ode:glfm:individual}. For prediction we use expression \eqref{eq:predictive:distribution} and 
the posterior distribution is found using expression \eqref{eq:posterior:latent:forces}, where the elements of the matrix
$\boldK_{\boldf,\boldu}$, $k_{f_d,u_q}(t,t')=k_{f^q_d,u_q}(t,t')$, are computed using 
\begin{align}\label{eq:cov:cross:ode:glfm:individual}
S_{d,q}\int_{\mathcal{T}}G_d(t-\tau)k_{u_q,u_q}(\tau,t')\dif\tau.
\end{align}

In section \ref{section:ode2}, we present in detail a second order latent force model and show
its application in the description of motion capture data.

\subsection{Multidimensional inputs}
In the sections above we have introduced latent force models for which the input variable is one-dimensional. 
For higher-dimensional inputs, $\mathbf{x}\in\mathbb{R}^p$, we can use linear partial differential equations to 
establish the dependence relationships between the latent forces and the outputs. The initial conditions turn into 
boundary conditions, 
specified by a set of functions that are linear combinations of $y_d(\mathbf{x})$ and its lower derivatives, evaluated
at a set of specific points of the input space. Inference and learning is done in a similar way to the one-input 
dimensional latent force model. Once the Green's function associated to the linear partial differential operator has been 
established, we employ similar equations to \eqref{eq:cov:ode:glfm:individual} and \eqref{eq:cov:cross:ode:glfm:individual}
to compute $k_{f_d,f_d'}(\mathbf{x},\mathbf{x}')$ and $k_{f_d,u_q}(\mathbf{x},\mathbf{x}')$ and the hyperparameters appearing
in the covariance function are estimated by maximizing the marginal likelihood. In section \ref{Sec:Diffusion}, we will 
present examples of latent force models with spatio-temporal inputs and a basic covariance with higher-dimensional inputs.

\subsection{Efficient approximations}
Learning the parameter vector $\bm{\theta}$ through the maximization of expression \eqref{eq:marginal:likelihood} 
involves the inversion of the matrix $\boldK_{\boldf,\boldf} + \bm{\Sigma}$, inversion that scales as $\mathcal{O}(D^3N^3)$.
For the single output case, this is $D=1$, different efficient approximations have been introduced in the machine learning
literature to reduce computational complexity including \citet{Csato:sparse00,Seeger:fast03,Quinonero:unifying05,
Snelson:pseudo05,Rasmussen:book06, Titsias:variational09}. Recently, \citet{Alvarez:sparse2009} introduced an efficient 
approximation for the case
$D>1$, which exploits the conditional independencies in equation \eqref{eq:convolution:operator}: assuming that only a few
number $K<N$ of values of $v(t)$ are known, then the set of outputs $f_d(t,v(t))$ are uniquely determined. The approximation
obtained shared characteristics with the Partially Independent Training Conditional (PITC) approximation introduced
in \citet{Quinonero:unifying05} and the authors of \citet{Alvarez:sparse2009} refer to the approximation as the PITC 
approximation for multiple-outputs. The set of values $\{v(t_k)\}_{k=1}^K$ are known as inducing variables, and 
the corresponding set of inputs, inducing inputs. This terminology has been  used before for the case in which $D=1$. 

A different type of approximation was presented in \citet{Alvarez:inducing10} based on variational methods. It is a 
generalization of \citet{Titsias:variational09} for multiple-output Gaussian processes. The approximation establishes a 
lower bound on the marginal likelihood and reduce computational complexity to $\mathcal{O}(DNK^2)$. The authors call this
approximation Deterministic Training Conditional Variational (DTCVAR) approximation for multiple-output GP regression, 
borrowing ideas from \citet{Quinonero:unifying05} and \citet{Titsias:variational09}.

\section{Second Order Dynamical System}\label{section:ode2}

In Section \ref{sec:intro} we introduced the analogy of a marionette's motion
being controlled by a reduced number of forces. Human motion capture
data consists of a skeleton and multivariate time courses of angles
which summarize the motion. This motion can be modelled with a set
of second order differential equations which, due to variations in
the centers of mass induced by the movement, are non-linear. The
simplification we consider for the latent force model is
to linearize these differential equations, resulting in the following
second order system,
\begin{align*}
M_d\frac{\text{d}^2 y_d(t)}{\text{d}t^2} + C_d \frac{\text{d}y_d(t)}{\text{d}t} + B_d y_d(t) & = 
\sum_{q=1}^{Q}{S_{d,q} u_q(t)}+\hat{e}_d(t).
\end{align*}
Whilst the above equation is not the correct physical model for our system, it will still be helpful 
when extrapolating predictions across different motions, as we shall see in the next section. Note also that, although 
similar to (\ref{Eq:GP-SIM:Differential_Equation}), the dynamic behavior of this system is much richer than that of the 
first order system, since it can exhibit inertia and resonance. In what follows, we will assume without loss of generality
that the masses are equal to one.

For the motion capture data $y_d(t)$ corresponds to a given observed angle over time, and its derivatives represent 
angular velocity and acceleration. The system is summarized by the undamped natural frequency, $\omega_{0d} =\sqrt{B_d}$, 
and the damping ratio, $\zeta_d=\frac{1}{2} C_d/\sqrt{B_d}$. Systems with a damping ratio greater than one are said to be 
overdamped, whereas underdamped systems exhibit resonance and have a damping ratio less than one. For critically damped 
systems $\zeta_d=1$, and finally, for undamped systems (i.e. no friction) $\zeta_d=0$.

Ignoring the initial conditions, the solution of the second order differential equation is given by the integral operator 
of equation \eqref{eq:convolution:operator}, with Green's function 
\begin{align}
 G_d(t,s) = \frac{1}{\omega_d}\exp(-\alpha_d(t-s))\sin(\omega_d(t-s)),
\label{Eq:GP-LFM:Greens}
\end{align}
where $\omega_d = \sqrt{4B_d-C_d^2}/2$ and $\alpha_d=C_d/2$. 

According to the general framework described in section \ref{subsection:hyperparameter:learning}, the covariance function
between the outputs is obtained by solving expression \eqref{eq:cov:ode:glfm:individual}, where $k_{u_q,u_q}(t,t')$ follows 
the SE form in equation \eqref{eq:cov:se}. Solution for $k_{f^q_d,f^q_{d'}}(t,t')$ is then given by \citep{Alvarez:lfm09} 
\begin{align*}
K_0\big[&h_q(\widetilde{\gamma}_{d'},\gamma_d,t,t') + h_q(\gamma_d,\widetilde{\gamma}_{d'},t',t) + 
\;h_q(\gamma_{d'},\widetilde{\gamma}_d,t,t') +\;h_q(\widetilde{\gamma}_d,\gamma_{d'},t',t) \\
- \; &h_q(\widetilde{\gamma}_{d'},\widetilde{\gamma}_d,t,t') - h_q(\widetilde{\gamma}_d,\widetilde{\gamma}_{d'},t',t) 
- \; h_q(\gamma_{d'},\gamma_d,t,t') - h_q(\gamma_d,\gamma_{d'},t',t)\big]
\end{align*}
where $K_0=\ell_q\sqrt{\pi}/8\omega_d\omega_{d'}$, $\gamma_d = \alpha_d + j \omega_d$ and 
$\widetilde{\gamma}_d = \alpha_d - j \omega_d$ and the functions $h_q(\widetilde{\gamma}_{d'},\gamma_d,t,t')$ follow 
\begin{align*}
h_q(\gamma_{d'},\gamma_d,t,t') &= \frac{\Upsilon_q(\gamma_{d'},t',t) - e^{-\gamma_d t} \Upsilon_q(\gamma_{d'},t',0)}{\gamma_d
+\gamma_{d'}},
\end{align*}
with 
\begin{align}
\Upsilon_{q}(\gamma_{d'},t,t')&=2e^{\left(\frac{\ell_q^2\gamma_{d'}^2}{4}\right)} e^{-\gamma_{d'}(t-t')} 
- e^{\left(-\frac{(t-t')^2}{\ell_q^2}\right)}\wfunc{j z_{d',q}(t)} - e^{\left(-\frac{(t')^2}{\ell_q^2}\right)}e^{(-\gamma_{d'} t)}
\wfunc{-j z_{d',q}(0)}, 
\label{Eq:GP-LFM:Upsilon_r_Kernel_Exp}
\end{align}
and $z_{d',q}(t) = (t-t')/\ell_q - (\ell_q \gamma_{d'})/2$. Note that $z_{d',q}(t) \in \Complex{}$, and
$\wfunc{jz}$ in \eqref{Eq:GP-LFM:Upsilon_r_Kernel_Exp}, for $z \in \Complex{}$, denotes Faddeeva's
function $\wfunc{jz} = \exp(z^2) \text{erfc}(z)$, where $\text{erfc}(z)$ is the complex version of the complementary
error function, $\text{erfc}(z) = 1 - \text{erf}(z) =
\frac{2}{\sqrt{\pi}} \int_{z}^{\infty}{\exp(-v^2) \text{d}v}$. Faddeeva's function is usually
considered the complex equivalent of the error function, since $|\wfunc{jz}|$ is bounded whenever
the imaginary part of $jz$ is greater or equal than zero, and is the key to achieving a good numerical
stability when computing \eqref{Eq:GP-LFM:Upsilon_r_Kernel_Exp} and its gradients.
 
Similarly, the cross-covariance between latent functions and outputs in equation \eqref{eq:cov:cross:ode:glfm:individual}
is given by
\[
k_{f_d^q, u_q}(t,t') = \frac{\ell_q S_{d,q} \sqrt{\pi}}{j 4 \omega_d}[\Upsilon_q(\widetilde{\gamma}_d,t,t') - 
\Upsilon_q(\gamma_d,t,t')],
\]

\subsubsection*{Motion Capture data}

Our motion capture data set is from the CMU motion capture data base.\footnote{The
CMU Graphics Lab Motion Capture Database was created with funding from NSF EIA-0196217 and
is available at \url{http://mocap.cs.cmu.edu}.} We considered two different types of movements: golf-swing and walking. 
For golf-swing we consider subject 64 motions 1, 2, 3 and 4,
and for walking we consider subject 35 motions 2 and 3; subject 10 motion 4; subject 12 motions 1, 2 and 3; subject 16, 
motions 15 and 21; subject 7 motions 1 and 2, and subject 8 motions 1 and 2. Subsequently, we will refer to the 
pair subject and motion by the notation $X(Y)$, where $X$ refers to the subject and $Y$ to the particular motion. 
The data was down-sampled by 4.\footnote{We selected specific frame intervals for each motion. For $64(1)$, 
frames $[120,400]$; for $64(2)$, frames $[170,420]$; for $64(3)$, frames $[100,300]$; and for $64(4)$, frames $[80,315]$. 
For $35(2)$, frames $[55,338]$; for $10(4)$, frames $[222,499]$; for $12(1)$, frames $[22,328]$; and for $16(15)$, frames 
$[62,342]$. For all other motions, we use all the frames.} 
Although each movement is described by time courses of 62 angles, we selected only the outputs whose 
signal-to-noise ratio was over 20 dB as explained in appendix \ref{appendix:prepro:mocap:data}, ending up with 
50 outputs for the golf-swing example and 33 outputs for the walking example.

We were interested in training on a subset of motions for each movement and testing on a different subset of motions for 
the same movement, to assess the model's ability to extrapolate. For testing, we condition on three angles associated to 
the root nodes and also on the first five and the last five output points of each other output. For the golf-swing, we 
use leave-one out cross-validation, in which one of the $64(Y)$ movements is left aside (with $Y=1,2,3$ or $4$) 
for testing, while we use the other three for training. For the walking example, 
we train using motions $35(2), 10(4), 12(1)$ and $16(15)$ and validate over all the other motions (8 in total). 

We use the above setup to train a LFM model with $Q=2$. We compare our model against MTGP and SLFM, also with $Q=2$. For
these three models, we use the DTCVAR efficient approximation with $K=30$ and fixed inducing-points placed equally 
spaced in the input interval. We also considered a regression model that directly predicts the angles of the body given 
the orientation of three root nodes using standard independent GPs with SE covariance functions. Results for all methods 
are summarized in Table \ref{table:cmu:results} in terms of root-mean-square error (RMSE) and percentage of explained 
variance (R$^2$). In the table, the measure shown is the mean of the measure in the validation set, plus and minus one
standard deviation.

\begin{table}[!ht]
\centering
\begin{tabular}{|c|c|c|c|}\hline 
\textbf{Movement} &  \textbf{Method} & \textbf{RMSE} &  \textbf{R$^2$ (\%)}\\\hline
\multirow{4}{*}{Golf swing} & IND GP & $21.55 \pm 2.35$  & $30.99\pm 9.67$\\
                         & MTGP  & $21.19\pm 2.18$ & $45.59\pm 7.86$\\
                         & SLFM  & $21.52\pm 1.93$ & $49.32\pm 3.03$\\
                         & LFM   & $\mathbf{18.09\pm 1.30}$ & $\mathbf{72.25\pm 3.08}$\\\hline
\multirow{4}{*}{Walking} & IND GP & $8.03\pm 2.55$ & $30.55\pm 10.64$\\
                         & MTGP & $7.75\pm 2.05$ & $37.77\pm 4.53$ \\
                         & SLFM & $7.81\pm 2.00$ & $36.84\pm 4.26$\\
                         & LFM  & $\mathbf{7.23\pm 2.18}$ & $\mathbf{48.15\pm 5.66}$\\\hline 
\end{tabular}
\caption{RMSE and R$^2$ for golf swing and walking}
\label{table:cmu:results}
\end{table}

We notice from table \ref{table:cmu:results} that the LFM outperforms the other methods both in terms of RMSE and R$^2$.
This is particularly true for the R$^2$ performance measure, indicating the ability that the LFM has for generating more
realistic motions.

\section{Partial Differential Equations and Latent Forces}
\label{Sec:Diffusion}

So far we have considered dynamical latent force models based on ordinary differential equations, leading to 
multioutput Gaussian processes which are functions of a single variable: time. As mentioned before, the methodology can 
also be applied in the context of partial differential equations to recover multioutput Gaussian processes 
which are functions of several inputs. We first show an example of spatio-temporal covariance obtained from the latent 
force model idea and then an example of a covariance function that, using a simplified version of the diffusion 
equation, allows an expression for higher-dimensional inputs. 

\subsection{Gap-gene network of Drosophila melanogaster}

In this section we show an example of a latent force model for a spatio-temporal domain. For illustration, we use 
gene expression data obtained from the Gap-gene network of the Drosophila melanogaster. We propose a linear model
that can account for the mechanistic behavior of the gene expression.

The gap gene network is responsible for the segmented body pattern of the \emph{Drosophila melanogaster}. During
the blastoderm stage of the development of the body, different maternal gradients determine the polarity of the embryo
along its anterior-posterior (A-P) axis \citep{Perkins:PLOS06}.

\begin{figure}[ht!]
\centering
\includegraphics[width=0.4\textwidth]{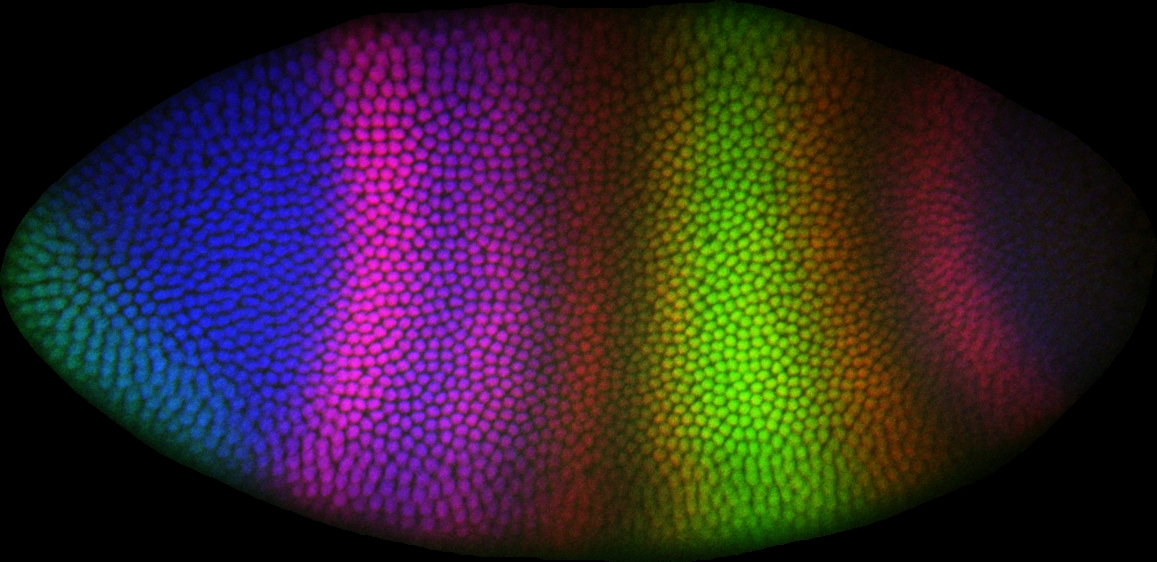}
\caption{Drosophila body segmentation genes. Blue stripes correspond to hunchback, green stripes to knirps and 
red stripes to eve-skipped at cleavage cycle 14A, temporal class 3.}\label{figure:drosSeg}
\end{figure}

Maternal gradient interact with the so called trunk gap genes, including \emph{hunchback (hb)}, \emph{Kr\"uppel (Kr)}, 
\emph{giant (gt)}, and \emph{knirps (kni)}, and this network of interactions establish the patterns of segmentation
of the Drosophila.

Figure \ref{figure:drosSeg} shows the gene expression of the hunchback, the knirps and the eve-skipped genes in a 
color-scale intensity image. The image corresponds to cleavage cycle 14A, temporal class 3.\footnote{The embryo name is 
dm12 and the image was taken from \url{http://urchin.spbcas.ru/flyex/}.}

The gap-gene network dynamics is usually represented using a set of coupled non-linear partial differential equations
\citep{Perkins:PLOS06,Gursky:PHYSICA04}
\begin{align*}
\frac{\partial y_d(x,t)}{\partial t}=\zeta(t) P_d(y(x,t)) - \lambda_d y_d(x,t)+\mathrm{D}_d\frac{\partial^2 y_d(x,t)}{\partial x^2},
\end{align*}
where $y_d(x,t)$ denotes the relative concentration of gap protein of the $d$-th gene at the space point $x$ and time point 
$t$. The term $P_d(y(x,t))$ accounts for production and it is a function, usually non-linear, of production of all other 
genes. The parameter $\lambda_d$ represents the decay and $\mathrm{D}_d$ the diffusion rate. The function $\zeta(t)$ accounts for 
changes occurring during the mitosis, in which the transcription is off \citep{Perkins:PLOS06}.

We linearize the equation above by replacing the non-linear term $\zeta(t)P_d(y(x,t))$ with the linear term 
$\sum_{q=1}^Q S_{d,q}u_q(x,t)$, where $S_{d,q}$ are sensitivities which account for the influence of the latent force 
$u_q(x,t)$ over the quantity of production of gene $d$. In this way, the new diffusion equation is given by
\begin{align}\notag
\frac{\partial y_d(x,t)}{\partial t}=\sum_{\forall q} S_{d,q}u_q(x,t)-\lambda_d y_d(x,t)+ \mathrm{D}_d
\frac{\partial^2 y_d(x,t)}{\partial x^2}.
\end{align}

This expression corresponds to a second order non-homogeneous partial differential equation. It is also parabolic with one 
space variable and constant coefficients. The exact solution of this equation is subject to particular initial and boundary 
conditions. For a first boundary value problem with domain $0\leq x\leq l$, initial condition $y_d(x,t=0)$ equal to zero,
and boundary conditions $y_d(x=0,t)$ and $y_d(x=l,t)$ both equal to zero, 
the solution to this equation is given by \citet{Polyanin:Handbook02, butkovskiy:pde1993,stakgold:green1998}
\begin{align}\notag
y_d(x,t)&=\sum_{q=1}^QS_{d,q}\int_0^t\int_0^lu_{q}(\xi,\tau)G_d(x,\xi,t-\tau)\dif{\xi}\dif{\tau},
\end{align}
where the Green's function $G_d(x,\xi,t)$ is given by
\begin{align}\notag
G_d(x,\xi,t)&=\frac{2}{l}e^{-\lambda_d t}\sum_{n=1}^\infty\sin\left(\frac{n\pi x}{l}\right)\sin
\left(\frac{n\pi \xi}{l}\right)e^{\left(-\frac{\mathrm{D}_d n^2\pi^2t}{l^2}\right)}.
\end{align}
We assume that the latent forces $u_{q}(x,t)$ follow a Gaussian process with covariance
function that factorizes across inputs dimensions, this is
\begin{align*}
k_{u_q,u_q}(x,t,x',t') = \exp\left(-\frac{(t-t')^ 2}{\left(\ell^t_q\right)^2}\right)\exp\left(-\frac{(x-x')^ 2}
{\left(\ell^x_q\right)^2}\right),
\end{align*}
where $\ell^t_q$ represents the length-scale along the time-input dimension and $\ell^x_q$ the length-scale along the space 
input dimension.
The covariance for the outputs $y_d(x,t)$, $k_{f_d^q, f_{d'}^q}(x,t,x',t')$, is computed using the expressions 
for the Green's function and the covariance of the latent forces, in a similar fashion to 
equation \eqref{eq:cov:ode:glfm:individual}, leading to 
\begin{align}\label{eq:dros:kernel}
k_{f_d^q, f_{d'}^q}(x,t,x',t')&=\frac{4}{\ell^2}\sum_{n=1}^{\infty}\sum_{m=1}^{\infty}k^{t}_{f_d^q, f_{d'}^q}(t,t')k^{x}_{f_d^q, f_{d'}^q}(x,x'),
\end{align}
where $k^{t}_{f_d^q, f_{d'}^q}(t,t')$ and $k^{x}_{f_d^q, f_{d'}^q}(x,x')$ are also kernel functions that depend on the indexes 
$n$ and $m$. The kernel function $k^{t}_{f_d^q, f_{d'}^q}(t,t')$ is given by expression \eqref{eq:cov:sim:basic} and 
$k^{x}_{f_d^q, f_{d'}^q}(x,x')$ is given by
\begin{align*}
k^{x}_{f_d^q, f_{d'}^q}(x,x')&=C(n,m,\ell_q^x)\sin(\omega_n x)\sin(\omega_m x'),
\end{align*}
where $\omega_n=\frac{n\pi}{\ell}$ and $\omega_m=\frac{m\pi}{\ell}$. The term $C(n,m,\ell_q^x)$ represents a function
that depends on the indexes $n$ and $m$, and on the length-scale of the space-input dimension. The expression for 
$C(n,m,\ell_q^x)$ is included in appendix \ref{appendix:cff}.

For completeness, we also include the cross-covariance between the outputs and the latent functions, which follows as
\begin{align*}
k_{f_d,u_q}(x,t,x',t')=\frac{2S_{d,q}}{l}\sum_{n=1}^{\infty}k^t_{f_d,u_q}(t,t')k^x_{f_d,u_q}(x,x'),
\end{align*}
where $k^t_{f_d,u_q}(t,t')$ is given by expression \eqref{eq:cross:output:latent}, and $k^x_{f_d,u_q}(x,x')$ follows
\begin{align*}
k^x_{f_d,u_q}(x,x')&=\sin\left(w_n x\right)C(x',n, \ell_q^x),
\end{align*}
where $C(x',n, \ell_q^x)$ is a function of $x'$, the index $n$ and the length-scale of the space input dimension. 
The expression for $C(x',n, \ell_q^x)$ is included in appendix \ref{appendix:cfu}.

\subsubsection*{Prediction of gene expression data}

We want to assess the contribution that a simple mechanistic assumption might bring to the prediction of gene expression 
data when compared to a covariance function that does not imply mechanistic assumptions. 

We refer to the covariance function obtained in the section before as the drosophila (DROS) kernel and compare against
the multi-task Gaussian process (MTGP) framework already mentioned in section \ref{section:lvm:to:lfm}. Covariance for 
the MTGP is a particular case of the latent force model covariance in equations \eqref{eq:cov:ode:glfm} and 
\eqref{eq:cov:ode:glfm:individual}. If we make $G_d(t-\tau)=\delta(t-\tau)$ in equation \eqref{eq:cov:ode:glfm}, and 
$k_{u_q,u_q}(t,t')=k_{u,u}(t,t')$ for all values of $q$, we get
\begin{align*}
k_{f_d,f_{d'}}(t,t')& = \sum_{q=1}^Q S_{d,q}S_{d',q}k_{u,u}(t,t').
\end{align*}
Our purpose is to compare the prediction performance of the covariance above and the DROS covariance function.

We use data from \citet{Perkins:PLOS06}, in particular, we have quantitative wild-type concentration profiles for the 
protein products of giant and knirps at 9 time points and 58 spatial locations. We work with a gene at a time and assume 
that the outputs correspond to the different
time points. This setup is very common in computer emulation of multivariate codes (see 
\citet{Conti:multi09,Rogers:towards08,Rougier:multi08}) in which the MTGP model is heavily used. For the DROS kernel,
we use 30 terms in each sum involved in its definition, in equation \eqref{eq:dros:kernel}.

We randomly select 20 spatial points for training the models, this is, for finding hyperparameters according to the 
description of subsection \ref{subsection:hyperparameter:learning}. The other 38 spatial points are used for validating 
the predictive performance. Results are shown in table \ref{table:dros:results} for five repetitions of the same 
experiment. It can be seen that the mechanistic assumption included in the GP model considerably outperforms 
a traditional approach like MTGP, for this particular task.

\begin{table}[!ht]
\centering
\begin{tabular}{|c|c|c|c|}\hline
\textbf{Gene} &  \textbf{Method} & \textbf{RMSE} &  \textbf{R$^2$ (\%)}\\ \hline
\multirow{2}{*}{giant}   & MTGP  & $ 26.56\pm 0.30 $ & $81.12\pm 0.01$\\
                         & DROS  &\hspace{0.5mm} $\mathbf{2.00\pm 0.35}$ &\hspace{-1.0mm} $\mathbf{99.78\pm 0.01}$\\\hline
\multirow{2}{*}{knirps}  & MTGP & $16.14\pm 8.44$ & $91.18\pm 2.77 $ \\
                         & DROS  & \hspace{1.5mm}$\mathbf{3.01\pm 0.81}$ & $\mathbf{99.60\pm 0.01}$\\\hline
\end{tabular}
\caption{RMSE and R$^2$ for protein data prediction}
\label{table:dros:results}
\end{table}

\subsection{Diffusion in the Swiss Jura}

The Jura data is a set of measurements of concentrations of several heavy metal pollutants collected from topsoil in a 
$14.5$ $\mbox{km}^2$ region of the Swiss Jura. We consider a latent function that represents how the pollutants were 
originally laid down. As time passes, we assume that the pollutants diffuse at different rates resulting in the 
concentrations observed in the data set. We use a simplified version of the heat equation of $p$ variables. The 
$p$-dimensional non-homogeneous heat equation is represented as
\begin{align*}
\frac{\partial \displacement_d(\mathbf{x},t)}{\partial t} &=\sum_{j=1}^{p}\kappa_{d,j}\frac{\partial^2
\displacement_d(\mathbf{x},t)}{\partial x_j^2}+\Phi(\boldx,t),
\end{align*}
where $p=2$ is the dimension of $\mathbf{x}$, the measured concentration of each
pollutant over space and time is given by $\displacement_d(\mathbf{x},t)$, $\kappa_{d,j}$ is the diffusion constant of output
$d$ in direction $p$, and $\Phi(\boldx,t)$ represents an external 
force, with $\boldx=\{x_j\}_{j=1}^p$. Assuming the domain $\mathbb{R}^p = \{-\infty<x_j<\infty;j=1,\ldots,p\}$ and initial 
condition prescribed by the set of latent forces,
\begin{align*}
u(\boldx)& = \sum_{q=1}^QS_{d,q}u_q(\boldx), \quad\text{at } t=0, 
\end{align*}
the solution to the system \citep{Polyanin:Handbook02} is then given by
\begin{align}
\displacement_d(\mathbf{x},t) &=\int_0^t\int_{\mathbb{R}^p}G_d(\mathbf{x},\mathbf{x}',t,\tau)\Phi(\mathbf{x}',\tau)
\dif\mathbf{x}'\dif{\tau}+\int_{\mathbb{R}^p}G_d(\mathbf{x},\mathbf{x}',t,0)\latentForce(\mathbf{x}')\dif\mathbf{x}',
\label{eq:sol:diffussion:d:variables:plus:phi}
\end{align}
where $G_q(\mathbf{x},\mathbf{x}',t,\tau)$ is the Green's function given by
\begin{align*}
G_d(\mathbf{x},\mathbf{x}',t,\tau) &=\frac{1}{2^p\pi^{p/2}\sqrt{\prod_{j=1}^pT_{d,j}}}\exp\left[-\sum_{j=1}^p\frac{(x_j-x'_j)^2}
{4T_{d,j}}\right],
\end{align*}
with $T_{d,j}(t,\tau)=\kappa_{d,j}(t-\tau)$. The covariance function we propose here is derived as follows. In equation  
\eqref{eq:sol:diffussion:d:variables:plus:phi}, we assume that the external force $\Phi(\boldx,t)$ is zero, following
\begin{align}
\displacement_d(\mathbf{x},t) &= \sum_{q=1}^QS_{d,q}\int_{\mathbb{R}^p}G_d(\mathbf{x},\mathbf{x}',t,0)\latentForce_q(\mathbf{x}')
\dif\mathbf{x}'.
\label{eq:sol:diffussion:d:variables}
\end{align}
We can write again the expression for the Green's function as
\begin{align*}
G_d(\mathbf{x},\mathbf{x}',t) &=\frac{1}{(2\pi)^{p/2}\sqrt{\prod_{j=1}^p2T_{d,j}}}\exp\left[-\sum_{j=1}^p\frac{(x_j-x'_j)^2}
{4T_{d,j}}\right]
=\frac{1}{(2\pi)^{p/2}\sqrt{\prod_{j=1}^p\ell_{d,j}}}\exp\left[-\sum_{j=1}^p\frac{(x_j-x'_j)^2}{2\ell_{d,j}}\right], 
\end{align*}
where $\ell_{d,j}=2T_{d,j}=2\kappa_{d,j}t$. The coefficient $\ell_{d,j}$ is a function of time. In our model for the diffusion
of the pollutant metals, we think of the data as a snapshot of the diffusion process. Consequently, we consider
the time instant of this snapshot as a parameter to be estimated. In other words, the measured concentration is given by
\begin{align}
\displacement_d(\mathbf{x}) &= \sum_{q=1}^QS_{d,q}\int_{\mathbb{R}^p}\widetilde{G}_d(\mathbf{x},\mathbf{x}')
\latentForce_q(\mathbf{x}')\dif\mathbf{x}',
\label{eq:sol:diffussion:d:variables:notime}
\end{align}
where $\widetilde{G}_d(\mathbf{x},\mathbf{x}')$ is the Green's function $G_d(\mathbf{x},\mathbf{x}',t)$ that considers
the variable $t$ as a parameter to be estimated through $\ell_{d,j}$. Expression for 
$\widetilde{G}_d(\mathbf{x},\mathbf{x}')$ corresponds to a Gaussian smoothing kernel, with diagonal covariance. This is
\begin{align*}
\widetilde{G}_d(\mathbf{x},\mathbf{x}')&=\frac{|\mathbf{P}_{d}|^{1/2}}{(2\pi)^{p/2}}
\exp\left[-\frac{1}{2}(\boldx-\boldx')^{\top}\mathbf{P}_{d} (\boldx-\boldx')\right],
\end{align*}
where $\mathbf{P}_d$ is a precision matrix, with diagonal form and entries 
$\{p_{d,j}=\frac{1}{\ell_{d,j}}\}_{j=1}^p$. 

If we take the latent function to be given by a GP with the Gaussian covariance function, we can 
compute the multiple output covariance functions analytically. The covariance function between the output functions, 
$k_{f_d^q,f_{d'}^q}(\mathbf{x},\mathbf{x}')$, is obtained as
\begin{align*}
&\frac{1}{(2\pi)^{p/2}|\mathbf{P}^q_{d,d'}|^{1/2}}
\exp\left[-\frac{1}{2}(\boldx-\boldx')^{\top}\left(\mathbf{P}^q_{d,d'}\right)^{-1} (\boldx-\boldx')\right],
\end{align*}
where $\mathbf{P}^q_{d,d'}=\mathbf{P}^{-1}_d+\mathbf{P}^{-1}_{d'}+\bm{\Lambda}_q^{-1}$, and $\bm{\Lambda}_q$ is the precision 
matrix associated to the Gaussian covariance of the latent force Gaussian process 
prior. The covariance function between the output and latent functions, $k_{f_d^q,u_q}(\mathbf{x},\mathbf{x}')$, is given by
\begin{align*}
&\frac{1}{(2\pi)^{p/2}|\mathbf{P}^q_{d}|^{1/2}}
\exp\left[-\frac{1}{2}(\boldx-\boldx')^{\top}\left(\mathbf{P}^q_{d}\right)^{-1} (\boldx-\boldx')\right],
\end{align*}
where $\mathbf{P}^q_{d}=\mathbf{P}^{-1}_d+\bm{\Lambda}_q^{-1}$.

\subsubsection*{Prediction of Metal Concentrations}

\begin{figure}[ht!]
\centering
\includegraphics[width=0.4\textwidth]{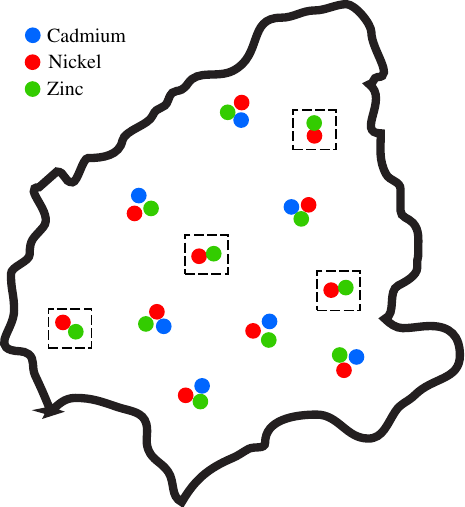}
\caption{Sketch of the topsoil of Swiss Jura. Secondary variables like nickel and zinc help in the prediction of the 
primary variable cadmium, in the squared-regions.}
\label{figure:juraTopSoil}
\end{figure}

We used our model to replicate the experiments described in \citet[pp.~248,249][]{Goovaerts:book97} in which a \emph{primary
  variable} (cadmium, cobalt, copper and lead) is predicted in conjunction with some \emph{secondary variables} 
(nickel and zinc for cadmium and cobalt; copper, nickel and zinc
for copper and lead).\footnote{Data available at \url{http://www.ai-geostats.org/}.} Figure \ref{figure:juraTopSoil} shows
an example of the prediction problem. For several sample locations we have access to the primary variable, for example 
cadmium, and the secondary variables, nickel and zinc. These sample locations are usually referred to as the 
\emph{prediction set}. At some other locations, we only have access to the secondary variables, as it is shown in the 
figure by the squared regions. In geostatistics, this configuration of sample locations is known as 
\emph{undersampled} or \emph{heterotopic}, where usually a few expensive measurements of the attribute of interest are 
supplemented by more abundant data on correlated attributes that are cheaper to sample.   

By conditioning on the values of the secondary variables at the prediction and validation sample locations, and 
the primary variables at the prediction sample locations, we can improve the prediction of the primary variables
at the validation locations. We compare results for the heat kernel with results from prediction using independent 
GPs for the metals, the multi-task Gaussian process and the semiparametric latent factor model. 
For our experiments we made use of ten repeats to report standard 
deviations. For each repeat, the data is divided into a different prediction set of $259$ locations and different 
validation set of $100$ locations. Root mean square errors and percentage of explained variance are shown in Tables 
\ref{table:jura:results:rmse} and \ref{table:jura:results:r2}, respectively.
\begin{table*}[!ht]
\centering
\begin{tabular}{|c|c|c|c|c|}\hline
\textbf{Method} & \textbf{Cadmium (Cd)} &  \textbf{Cobalt (Co)}& \textbf{Copper (Cu)} &  \textbf{Lead (Pb)}\\\hline 
IND GP &  $0.8353\pm 0.0898$ & $2.2997\pm 0.1388 $& $18.9616\pm 3.4404 $ & $28.1768\pm 5.8005$\\
MTGP~~ ($Q=1$)&  $0.7638\pm 0.1016$ & $2.2892\pm 0.1792 $ & $14.4179\pm 2.7119$ & $21.5861\pm 4.1888 $\\
HEATK~~($Q=1$)& $0.6773\pm 0.0628$ & \hspace{3mm}$2.06\pm 0.0887$& $13.1788\pm 2.6446$ & $17.9839\pm 2.9450$\\
MTGP~~ ($Q=2$)&   $0.6980\pm 0.0832$ & $2.1299\pm 0.1983 $ & $12.7340\pm 2.2104$ & $17.9399\pm 1.9981 $\\
SLFM~~~($Q=2$)&  $0.6941\pm 0.0834$ & \hspace{1.5mm}$2.172\pm 0.1204$& $12.8935\pm 2.6125$ & $17.9024\pm 2.0966$\\
HEATK~~($Q=2$)& $\mathbf{0.6759\pm0.0623}$ & $\mathbf{2.0345\pm 0.0943}$& $\mathbf{12.5971\pm2.4842}$ 
& $\mathbf{17.5571\pm 2.6076}$\\\hline
\end{tabular}
\caption{RMSE for pollutant metal prediction}
\label{table:jura:results:rmse}
\end{table*}

\begin{table*}[!ht]
\centering
\begin{tabular}{|c|c|c|c|c|}\hline
\textbf{Method} & \textbf{Cadmium (Cd)} &  \textbf{Cobalt (Co)}& \textbf{Copper (Cu)} &  \textbf{Lead (Pb)}\\\hline
IND GP & $15.07\pm 7.43 $ & $57.81\pm 7.19$& $25.84\pm 7.54 $ & \hspace{1.5mm}$23.48\pm 10.40$\\
MTGP~~($Q=1$)  & $27.25\pm 5.89$ & $58.45\pm 5.71 $ & $58.84\pm 8.35$ & \hspace{1.5mm}$56.85\pm 11.60 $\\
HEATK~~($Q=1$)  & $43.83\pm8.71$ & $66.19\pm 4.60$& $65.55\pm8.21$ & $\mathbf{71.45\pm 5.78}$\\
MTGP~~($Q=2$)  & $40.30\pm 5.17$ & $64.13\pm 5.10$ & $67.51\pm 8.36$ & $69.70\pm 6.90 $\\
SLFM~~($Q=2$)  & $40.97\pm 5.15$ & $62.49\pm 5.41$& $67.35\pm 8.29$ & $70.21\pm 6.04$\\
HEATK~~($Q=2$)& $ \mathbf{43.94\pm 6.56} $ & $\mathbf{67.17\pm4.30}$& $\mathbf{68.40 \pm 6.46}$ & $70.55\pm6.88$\\\hline
\end{tabular}
\caption{R$^2$ for pollutant metal prediction}
\label{table:jura:results:r2}
\end{table*}

Note from both tables that all methods outperform independent Gaussian processes, in terms of RMSE and 
explained variance. For one latent function ($Q=1$), the Gaussian process with Heat kernel render better results than 
multi-task GPs (in this case, the multi-task GP is equivalent to the semiparametric latent factor model). However, when 
increasing the value of the latent forces to two 
($Q=2$), performances for all methods are quite similar. There is a still a gain in performance when using the Heat
kernel, although the results are within the standard deviation. Also, when comparing the performances for the GP with Heat 
kernel using one and two latent forces, we notice that both measures are quite similar. In summary, the heat kernel 
provides a simplified explanation for the outputs, in the sense that, using only one latent force, we provide better 
performances in terms of RMSE and explained variance.

\section{Related work}\label{section:related:work}

Differential equations are the cornerstone in a diverse range of engineering fields and applied sciences. However, their use
for inference in  statistics and machine learning has been less studied. The main field in which they have been used is 
known as \emph{functional data analysis} \citep{Ramsay:functionalData2005}. 

From the frequentist statistics point of view, the literature in functional data analysis has been concerned  
with the problem of parameter estimation in differential equations \citep{Poyton:PDA:2006, Ramsay:PDANonlinear:2007}: 
given a differential equation with unknown coefficients $\{\boldA_m\}_{m=0}^M$, how do we use data to fit those parameters? 
Notice that there is a subtle difference between those techniques and the latent force model. While these parameter
estimation methods start with a very accurate description of the interactions in the system via the differential equation
(the differential equation might even be non-linear \citep{Perkins:PLOS06}),
in the latent force model, we use the differential equation as part of the modeling problem: the differential equation is 
used as a way to introduce prior knowledge over a system for which we do not know the real dynamics, but for which we hope
some important features of that dynamics could be expressed. Having said that, we review some of the parameter estimation 
methods because they also deal with differential equations with an uncertainty background.
 
Classical approaches to fit parameters $\bm{\theta}$ of differential equations to observed data include numerical 
approximations of initial value problems and collocation methods (references \citet{Ramsay:PDANonlinear:2007} and
\citet{Brewer:odeTimeCourseData:2008} provide reviews and detailed descriptions of additional methods). 

The solution by numerical approximations include an iterative process in which given an initial set of parameters 
$\bm{\theta}_0$ and a set of initial conditions $\mathbf{y}_0$, a numerical
method is used to solve the differential equation. The parameters of the differential equation are then optimized by 
minimizing an error criterion between the approximated solution and the observed data. For exposition, we assume in 
equation \eqref{eq:lfm:higher:order:individual} that $D=1$, $Q=1$ and $S_{1,1}=1$. We are
interested in finding the solution $y(t)$ to the following differential equation, with unknown parameters 
$\params=\{A_m\}_{m=0}^M$,
\begin{align*}
\mathcal{D}^{M}_0\displacement(t) &=\sum_{m=0}^MA_m\mathcal{D}^{m}\displacement(t)= u(t),
\end{align*}

In the classical approach, we assume that we have access to a vector of initial conditions, 
$\boldy_0$ and data for $u(t)$, $\boldu$. We start with an initial guess for the parameter vector $\params_0$ and solve 
numerically the differential equation to find a solution $\widetilde{\boldy}$. An updated parameter vector 
$\widetilde{\params}$ is obtained by minimizing   
\begin{align*}
E(\params)&=\sum_{n=1}^N\|\widetilde{y}(t_n) - y(t_n)\|,
\end{align*} 
through any gradient descent method. To use any of those methods, we must be able to compute 
$\partial E(\params)/\partial \params$, which is equivalent to compute $\partial y(t)/\partial \params$. In general, when
we do not have access to $\partial y(t)/\partial \params$, we can compute it using what is known as the \emph{sensitivity 
equations} (see \citet{Bard:NonlinearPE:1974}, chapter 8, for detailed explanations), which are solved along with the ODE 
equation that provides the partial solution $\widetilde{\boldy}$. 
Once a new parameter vector $\widetilde{\params}$ has been found, 
the same steps are repeated until some convergence criterion is 
satisfied. If the initial conditions are not available, they can be considered as additional elements of the parameter
vector $\params$ and optimized in the same gradient descent method. 

In collocation methods, the solution of the differential equation is approximated using a set of basis functions, 
$\{\phi_i(t)\}_{i=1}^J$, this is $y(t)=\sum_{i=1}^J\beta_{i}\phi_i(t)$. The basis functions must be sufficiently smooth so 
that the derivatives of the unknown function, appearing in the differential equation, can be obtained by differentiation 
of the basis representation of the solution, this is, $\mathcal{D}^my(t)=\sum\beta_i\mathcal{D}^m\phi_i(t)$. Collocation 
methods also use an iterative procedure for fitting the additional parameters involved in the differential equation. 
Once the solution and its derivatives have been approximated using the set of basis functions, minimization of an error 
criteria is used to
estimate the parameters of the differential equation. Principal differential analysis (PDA) \citep{Ramsay:PDA:1996} is one 
example of a collocation method in which the basis functions are \emph{splines}. In PDA, the parameters of the differential 
equation are obtained by minimizing the squared residuals of the higher order derivative 
$\mathcal{D}^My(t)$ and the weighted sum of derivatives $\{\mathcal{D}^my(t)\}_{m=0}^{M-1}$, instead of the squared residuals
between the approximated solution and the observed data. \\

An example of a collocation method augmented with Gaussian process priors was introduced by 
\citet{Graepel:noisyLinearOperator:2003}. Graepel starts with noisy observations,
$\widehat{y}(t)$, of the differential equation $\mathcal{D}^M_0y(t)$, such that $\widehat{y}(t)
\sim\gauss(\mathcal{D}^M_0y(t), \sigma_y)$. The solution $y(t)$ is expressed using a basis representation, 
$y(t)=\sum\beta_i\phi_i(t)$. A Gaussian prior is placed over $\bm{\beta}=[\beta_1, \ldots, \beta_J]$, and its posterior 
computed under the above likelihood. With the posterior over $\bm{\beta}$, the predictive distribution for 
$\widehat{y}(t_*)$ can be readily computed, being a function of the matrix $\mathcal{D}^M_0\bm{\Phi}$ with elements 
$\{\mathcal{D}_0^M\phi_i(t_n)\}_{n=1,i=1}^{N,J}$. It turns out that products 
$\mathcal{D}^M_0\bm{\Phi}\big(\mathcal{D}^M_0\bm{\Phi}\big)^{\top}$ that appear in this predictive distribution 
have individual elements that can be written using the sum 
$\sum_{i=1}^J\mathcal{D}_0^M\phi_i(t_n)\mathcal{D}_0^M\phi_i(t_{n'})=\mathcal{D}_{0,t}^M\mathcal{D}_{0,t'}^M
\sum_{i=1}^J\phi_i(t_n)\phi_i(t_{n'})$ or, using a kernel representation for
the inner products $k(t_n,t_{n'})=\sum_{i=1}^J\phi_i(t_n)\phi_i(t_{n'})$, as 
$k_{\mathcal{D}_{0,t}^M,\mathcal{D}_{0,t'}^M}(t_n,t_{n'})$, where this covariance is obtained by taking $\mathcal{D}^M_0$ 
derivatives of $k(t,t')$ with respect to $t$ and $\mathcal{D}^M_0$ derivatives with respect to $t'$. In other words, the 
result of the differential equation $\mathcal{D}^M_0y(t)$ is assumed to follow a Gaussian process prior with covariance
$k_{\mathcal{D}_{0,t}^M,\mathcal{D}_{0,t'}^M}(t,t')$. An approximated solution $\widetilde{y}(t)$ can be computed through the
expansion $\widetilde{y}(t)=\sum_{n=1}^N\alpha_{n}k_{\mathcal{D}_{0,t'}^M}(t,t_n)$, where $\alpha_n$ is an element of the vector
$(\boldK_{\mathcal{D}_{0,t}^M,\mathcal{D}_{0,t'}^M}+\sigma_y\eye_N)^{-1}\widehat{\boldy}$, where 
$\boldK_{\mathcal{D}_{0,t}^M,\mathcal{D}_{0,t'}^M}$ is a matrix with entries $k_{\mathcal{D}_{0,t}^M,\mathcal{D}_{0,t'}^M}(t_n,t_{n'})$
and $\widehat{\boldy}$ are noisy observations of $\mathcal{D}^M_0y(t)$.\\

Although, we presented the above methods in the context of linear ODEs, solutions by numerical approximations and 
collocation methods are applied to non-linear ODEs as well.\\

Gaussian processes have been used as models for systems identification \citep{Solak:derivativesGP:2003, 
Kocijan:identificationGPs:2005, Calderhead:mcmcNonODE:2008, Thompson:phdThesis:2009}. In 
\citet{Solak:derivativesGP:2003}, a non-linear dynamical system is linearized around an 
equilibrium point by means of a Taylor series expansion \citep{Thompson:phdThesis:2009}, 
\begin{align*}
y(t) = \sum_{j=0}^{\infty}\frac{y^{(j)}(a)}{j!}(t-a)^j,
\end{align*}
with $a$ the equilibrium point. For a finite value of terms, the linearization above can be seen as a regression problem 
in which the covariates correspond to the terms $(t-a)^j$ and the derivatives $y^{(j)}(a)$ as regression coefficients. The 
derivatives are assumed to follow a Gaussian process prior with a covariance function that is obtained as 
$k^{(j,j')}(t,t')$, where the superscript $j$ indicates how many derivative of $k(t,t')$ are taken with respect to $t$ and 
the superscript $j'$ indicates how many derivatives of $k(t,t')$ are taken with respect to $t'$. Derivatives are then 
estimated a posteriori through standard Bayesian linear regression. An important consequence of including derivative 
information in the inference process is that the 
uncertainty in the posterior prediction is reduced as compared to using only function observations. This aspect of 
derivative information have been exploited in the theory of computer emulation to reduce the uncertainty in experimental 
design problems \citep{Morris:computerExperiments:derivatives:1993, Mitchell:computerExperiments:derivatives:1994}.

Gaussian processes have also been used to model the output $y(t)$ at time $t_k$ as a function of its $L$ 
previous samples $\{y(t-t_{k-l})\}_{l=1}^L$, a common setup in the classical theory of systems identification 
\citep{Ljung:system:id:1999}. The particular dependency $y(t)=g(\{y(t-t_{k-l})\}_{l=1}^L)$, where $g(\cdot)$ is a general non-
linear function, is modelled using a Gaussian
process prior and the predicted value for the output $y_*(t_k)$ is used as a new input for multi-step ahead prediction 
at times $t_{j}$, with $j>k$ \citep{Kocijan:identificationGPs:2005}. Uncertainty about $y_*(t_k)$ can also be incorporated  
for predictions of future output values \citep{Girard:multipleStepAhead:2003}.      

On the other hand, multivariate systems with Gaussian process priors have been thoroughly studied in the spatial 
analysis and geostatistics literature \citep{Higdon:convolutions02, Boyle:dependent04, Journel:miningBook78, 
Cressie:spatialdataBook:1993, Goovaerts:book97, Wackernagel:book03}. In short, a valid covariance function for multi-output
processes can be generated using the linear model of coregionalization (LMC). In the LMC, each output $y_d(t)$ is 
represented as a linear combination of a series of basic processes $\{u_q\}_{q=1}^Q$, some of which share the same 
covariance function $k_{u_q,u_q}(t,t')$. Both, the semiparametric latent factor model \citep{Teh:semiparametric05} and the 
multi-task GP \citep{Bonilla:multi07} can be seen as 
particular cases of the LMC \citep{Alvarez:kernelVectorial:report:2011}. \citet{Higdon:convolutions02} proposed the direct 
use of a expression 
\eqref{eq:convolution:operator} to obtain a valid covariance function for multiple outputs and referred to this kind of 
construction as process convolutions. Process convolutions for constructing covariances for single output GP 
had already been proposed by \citet{Barry:balckbox96,verHoef:convolution98}. 
\citet{Calder:convolution07} reviews several extensions of the single process convolution covariance. 
\citet{Boyle:dependent04} introduced the process convolution idea for multiple outputs to
the machine learning audience. \citet{Boyle:phdThesis:2007} developed the idea of using impulse responses of filters
to represent $G_d(t,s)$, assuming the process $v(t)$ was white Gaussian noise. Independently, 
\citet{RMurray:transformationGPs:2005} also introduced the idea of transforming a Gaussian process prior using a 
discretized version of the integral operator of equation \eqref{eq:convolution:operator}. Such transformation could be 
applied for the purposes of fusing the information from multiple sensors (a similar setup to the latent force model but
with a discretized convolution), for solving inverse problems in reconstruction of images or for reducing 
computational complexity working with the filtered data in the transformed space \citep{Shi:LargeData:FilteredGPs:2005}.    

There has been a recent interest in introducing Gaussian processes in the state space 
formulation of dynamical systems \citep{Ko:gpukf:2007, Deisenroth:AnalyticModelBasedGP:2009, Turner:state:space:2010} for 
the representation of the possible nonlinear relationships between the latent space and between the latent space and the 
observation space. Going back to the formulation of the dimensionality reduction model, we have
\begin{align*}
\boldu_{t_n}=\mathbf{g}_1(\boldu_{t_{n-1}})+\bm{\eta},\\
\boldy_{t_n}=\mathbf{g}_2(\boldu_{t_n})+\bm{\epsilon},
\end{align*}
where $\bm{\eta}$ and $\bm{\xi}$ are noise processes and $\mathbf{g}_1(\cdot)$ and $\mathbf{g}_2(\cdot)$ are general 
non-linear functions. Usually $\mathbf{g}_1(\cdot)$ and $\mathbf{g}_2(\cdot)$ are unknown, and research on this area has 
focused on developing a practical framework for inference when assigning Gaussian process priors to both functions. 

Finally, it is important to highlight the work of Calder 
\citep{Calder:thesis03,Calder:kalmanConvolution07, Calder:randomWalks08} as an alternative
to multiple-output modeling. Her work can be seen in the context of state-space models, 
\begin{align*}
\boldu_{t_n}=\boldu_{t_{n-1}}+\bm{\eta},\\
\boldy_{t_n}=\mathbf{G}_{t_n}\boldu_{t_n}+\bm{\epsilon},
\end{align*}
where $\boldy_{t_n}$ and $\boldu_{t_n}$ are related through a discrete convolution over an independent spatial variable. 
This is, for a fixed $t_n$, $y^{t_n}_{d}(\bolds)=\sum_q\sum_i G^{t_n}_d(\bolds - \mathbf{z}_i)u_q^{t_n}(\mathbf{z}_i)$ for a 
grid of $I$ spatial inputs $\{\mathbf{z}_i\}_{i=1}^I$.

\section{Conclusion}\label{section:conclusion}

In this paper we have presented a hybrid approach to modelling that
sits between a fully mechanistic and a data driven approach. We used
Gaussian process priors and linear differential equations to model
interactions between different variables. The result is the
formulation of a probabilistic model, based on a kernel function, that
encodes the coupled behavior of several dynamical systems and allows
for more accurate predictions.  The implementation of latent force
models introduced in this paper can be extended in several ways,
including:\\
\emph{Non-linear Latent Force Models.} If the likelihood function is
not Gaussian the inference process has to be accomplished in a
different way, through a Laplace approximation \citep{Lawrence:gpsim2007a} or sampling
techniques \citep{Titsias:control:vars:2009}.\\
\emph{Cascaded Latent Force Models.} For the above presentation of the latent force model, we assumed that the covariances
$k_{u_q,u_q}(t,t')$ were squared-exponential. However, more structured
covariances can be used. For example, in \citet{Honkela:PNAS10}, the authors use a cascaded system to
describe gene expression data for which a first order system, like the
one presented in subsection \ref{subsection:first:lfm}, has as
inputs $u_q(t)$ governed by Gaussian processes with covariance function \eqref{eq:cov:sim:basic}.\\
\emph{Stochastic Latent Force Models.} If the latent forces $u_q(t)$ are white noise processes, then 
the corresponding differential equations are stochastic, and the covariances obtained in such cases lead
to stochastic latent force models. In \citet{Alvarez:inducing10}, a first-order stochastic latent force model is 
employed for describing the behavior of a multivariate financial dataset: the foreign exchange rate with respect to the 
dollar of ten of the top international currencies and three precious metals.\\
\emph{Switching dynamical Latent Force Models.} A further extension of
the LFM framework allows the parameter vector $\bm{\theta}$ to have
discrete changes as function of the input time. In
\citet{Alvarez:switched11} this model was used for the segmentation of
movements performed by a Barrett WAM robot as haptic input device.

\section*{Acknowledgments}

DL has been partly financed by Comunidad de Madrid (project PRO-MULTIDIS-CM, S-0505/TIC/0233), and by the Spanish 
government (CICYT project TEC2006-13514-C02-01 and researh grant JC2008-00219). MA and NL have been financed by a 
Google Research Award and EPSRC Grant No EP/F005687/1 ``Gaussian Processes for Systems Identification with Applications in 
Systems Biology''. MA also acknowledges the support from the Overseas Research Student Award Scheme (ORSAS), 
from the School of Computer Science of the University of Manchester and from the Universidad Tecnológica de Pereira, 
Colombia.

\appendix

\section{Preprocessing for the mocap data}\label{appendix:prepro:mocap:data}

For selecting the subset of angles for each of the motions for the
golf-swing movement and the walking movement, we use as performance
measure the signal-to-noise ratio obtained in the following way. We
train a GP regressor for each output, employing a covariance function
that is the sum of a squared exponential kernel and a white Gaussian
noise,
\begin{align*}
\sigma^2_S\exp\left[-\frac{(\boldx-\boldx')^2}{2\ell^2}\right] + \sigma^2_N\delta(\boldx,\boldx'),
\end{align*}
where $\sigma^2_S$ and $\sigma^2_N$ are variance parameters, and $\delta(\boldx,\boldx')$ is the Dirac delta function.
For each output, we compute the signal-to-noise ratio as $10\log_{10}(\sigma^2_S/\sigma^2_N)$.

\section{Expression for $C(n,m,\ell_x)$}\label{appendix:cff}
For simplicity, we assume $Q=1$ and write $\ell_q^x=\ell_x$. The expression for $C(n,m,\ell_x)$ is given by
\begin{align*}
C(n,m,\ell_x)&=\int_0^l\int_0^l\sin\left(w_n\xi\right)
\sin\left(w_m\xi'\right)e^{\left[-\frac{\left(\xi-\xi'\right)^2}{\ell^2_x}\right]}\dif\xi'\dif\xi.
\end{align*}
The solution of this double integral depends upon the relative values of $n$ and $m$. If $n\ne m$, and $n$ and $m$ are 
both even or both odd, then the analytical expression for $C(n,m,\ell_x)$ is 
\begin{align*}
\left(\frac{\ell_x l}{\sqrt{\pi}(m^2-n^2)}\right)\left\{n\mathcal{I}\left[\mathcal{W}(m,\ell_x)\right]-\;m\mathcal{I}\left[\mathcal{W}(n,\ell_x)\right]\right\},
\end{align*}
where $\mathcal{I}[\cdot]$ is an operator that takes the imaginary part of the argument and $\mathcal{W}(m,\ell_x)$ 
is given by
\begin{align*}
\mathcal{W}(m,\ell_x)&=\wfunc{jz_1^{\gamma_m}}-e^{-\left(\frac{l}{\ell_x}\right)^2}e^{-\gamma_m l}\wfunc{jz_2^{\gamma_m}},
\end{align*}
being $z_1^{\gamma_m}=\frac{\ell_x \gamma_m}{2}$, $z_2^{\gamma_m}=\frac{l}{\ell_x}+\frac{\ell_x \gamma_m}{2}$ and 
$\gamma_m=j\omega_m$.\\

The term $C(n,m,\ell_x)$ is zero if, for $n\ne m$, $n$ is even and $m$ is odd or viceversa.\\

Finally, when $n=m$, the expression for $C(n,n,\ell_x)$ follows as 
\begin{align*}
\frac{\ell_x\sqrt{\pi}\,l}{2}\bigg\{\mathcal{R}\big[\mathcal{W}&(n,\ell_x)\big]
-\mathcal{I}\left[\mathcal{W}(n,\ell_x)\right]\Big[\frac{\ell_x^2n\pi}{2l^2}+\frac{1}{n\pi}\Big]\bigg\}
+\frac{\ell^2_x}{2}\left[e^{-(\frac{l}{\ell_x})^2}\cos(n\pi)-1\right],
\end{align*}
where $\mathcal{R}[\cdot]$ is an operator that takes the real part of the argument.

\section{Expression for $C(x',n,\ell_x)$}\label{appendix:cfu}

As in appendix \ref{appendix:cff}, we assume $Q=1$ and $\ell_q^x=\ell_x$. The expression $C(x',n,\ell_x)$ is as
\begin{align*}
\frac{\ell_x\sqrt{\pi}}{2}\mathcal{I}\left[e^{-\left(\frac{x'-l}{\ell_x}\right)^2}e^{\gamma_nl}
\wfunc{jz_2^{\gamma_n,x'}}-e^{-\left(\frac{x'}{\ell_x}\right)^2}\wfunc{jz_1^{\gamma_n,x'}}\right],
\end{align*}
with $z_1^{\gamma_n,x'}=\frac{x'}{\ell_x}+\frac{\ell_x \gamma_n}{2}$, $z_2^{\gamma_n,x'}=\frac{x'-l}{\ell_x}
+\frac{\ell_x\gamma_n}{2}$, $\gamma_n = j\omega_n$ and $\mathcal{I}[\cdot]$ is an operator that takes the imaginary part 
of the argument.

\bibliographystyle{plainnat}
\bibliography{../../sparseMultigp/reportbiblio}

\end{document}